\documentclass[letterpaper]{article} 
\usepackage{aaai25}  
\usepackage{times}  
\usepackage{helvet}  
\usepackage{courier}  
\usepackage[hyphens]{url}  
\usepackage{graphicx} 
\usepackage{natbib}  
\usepackage{caption} 
\usepackage[table]{xcolor}

\usepackage{color}
\usepackage{booktabs} 
 \usepackage{multirow}
\usepackage{colortbl}

\usepackage{algorithm}
\usepackage{algorithmic}
\usepackage{amsfonts}
\usepackage{amsmath}
\newcounter{checksubsection}
\newcounter{checkitem}[checksubsection]

\newcommand{\checksubsection}[1]{%
  \refstepcounter{checksubsection}%
  \paragraph{\arabic{checksubsection}. #1}%
  \setcounter{checkitem}{0}%
}

\newcommand{\checkitem}{%
  \refstepcounter{checkitem}%
  \item[\arabic{checksubsection}.\arabic{checkitem}.]%
}
\newcommand{\question}[2]{\normalcolor\checkitem #1 #2 \color{blue}}
\newcommand{\ifyespoints}[1]{\makebox[0pt][l]{\hspace{-15pt}\normalcolor #1}}

\usepackage{newfloat}
\usepackage{listings}
\DeclareCaptionStyle{ruled}{labelfont=normalfont,labelsep=colon,strut=off} 
\floatstyle{ruled}
\newfloat{listing}{tb}{lst}{}
\floatname{listing}{Listing}
\title{Progressive Flow-inspired Unfolding for Spectral Compressive Imaging}
\author{
    Xiaodong Wang\textsuperscript{\rm 1, 2}, Ping Wang\textsuperscript{\rm 1,2 }, Zijun He\textsuperscript{\rm 1,2}, Mengjie Qin\textsuperscript{\rm 2}, Xin Yuan\textsuperscript{\rm 2 \thanks{Corresponding author.}}
}
\affiliations{
    \textsuperscript{\rm 1} Zhejiang University, Hangzhou, China\\
    \textsuperscript{\rm 2} Westlake University, Hangzhou, China \\
    
    \{wangxiaodong, wangping, hezijun, qinmengjie, xyuan\}@westlake.edu.cn
}

\usepackage{bibentry}

\begin{document}

\maketitle

\begin{abstract}
Coded aperture snapshot spectral imaging (CASSI) retrieves a 3D hyperspectral image (HSI) from a single 2D compressed measurement, which is a highly challenging reconstruction task. Recent deep unfolding networks (DUNs), empowered by explicit data-fidelity updates and implicit deep denoisers, have achieved the state of the art in CASSI reconstruction. However, existing unfolding approaches suffer from uncontrollable reconstruction trajectories, leading to abrupt quality jumps and non-gradual refinement across stages. Inspired by diffusion trajectories and flow matching, we propose a novel trajectory-controllable unfolding framework that enforces smooth, continuous optimization paths from noisy initial estimates to high-quality reconstructions. To achieve computational efficiency, we design an efficient spatial-spectral Transformer tailored for hyperspectral reconstruction, along with a frequency-domain fusion module to gurantee feature consistency. Experiments on simulation and real data demonstrate that our method achieves better reconstruction quality and efficiency than prior state-of-the-art approaches.
\end{abstract}

%

\section{Introduction}

Hyperspectral imaging (HSI) captures fine-grained spectral signatures and has been widely applied in remote sensing, biomedical imaging, and material analysis. However, acquiring full 3D hyperspectral cubes is time-consuming and hardware-demanding. To address this, Coded Aperture Snapshot Spectral Imaging (CASSI) ~\cite{arce2013compressive, yuan2021snapshot} has emerged as a promising solution that captures the full spectral volume in a single shot by encoding spatial-spectral information through a coded aperture and disperser.

Reconstructing the hyperspectral image from the CASSI measurement is a severely ill-posed inverse problem. Consider the recovery of the target hyperspectral cube $\mathbf{x} \in \mathbb{R}^n$ from the compressed measurements $\mathbf{y} \in \mathbb{R}^m$ obtained through
\begin{equation}
\mathbf{y} = \mathbf{A}\mathbf{x} + \mathbf{n},
\label{eq:cassi_model}
\end{equation}
where $\mathbf{A} \in \mathbb{R}^{m \times n}$ with $m \ll n$ encodes the CASSI system, and $\mathbf{n}$ denotes additive noise. This severe ill-posed problem is typically addressed via maximum a posteriori (MAP) estimation~\cite{kamilov2023plug,venkatakrishnan2013plug}:
\begin{equation}
\hat{\mathbf{x}} = \arg\min_{\mathbf{x}} \frac{1}{2}\|\mathbf{y} - \mathbf{A}\mathbf{x}\|_2^2 - \log p(\mathbf{x}),
\label{eq:map_formulation}
\end{equation}
where $p(\mathbf{x})$ is a \textit{prior} for the distribution of natural images. Here, the data fidelity term $\frac{1}{2}\|\mathbf{y} - \mathbf{A}\mathbf{x}\|_2^2$ corresponds to $-\log p(\mathbf{y}|\mathbf{x})$ under Gaussian noise assumption. Solving this optimization typically alternates between gradient updates for data consistency and proximal operations that enforce the prior $p(\mathbf{x})$. While the data consistency update has a closed-form solution, the challenge lies in designing appropriate priors that capture the complex spatial-spectral correlations inherent in hyperspectral data. 

Based on different prior modeling strategies, reconstruction methods are divided into four categories. Iterative optimization methods solve Eq.~\eqref{eq:map_formulation} using hand-crafted priors such as sparsity~\cite{yuan2016generalized, chen2023prior}, and low-rank constraints~\cite{gelvez2020nonlocal}, offering theoretical guarantees but limited expressiveness and requiring extensive hyperparameter tuning. Plug-and-play (PnP) methods~\cite{qiu2021effective, zheng2021deep} replace the prior term in iterative solvers with pre-trained minimum mean square error (MMSE) denoisers, although most available denoisers are trained on RGB/grayscale images, thus lacking the capability to model complex spectral correlations. Deep unfolding networks~\cite{cai2022degradation,dong2023residual,li2023pixel,wu2024latent,qin2025detail,zhang2024improving,zhang2024dual} parameterize each iteration of prior updating as learnable neural network, enabling data-driven
optimization while maintaining algorithmic interpretability to achieve state-of-the-art performance. End-to-end networks~\cite{wang2025sˆ, meng2020end, cai2022mask} abandon explicit optimization structures entirely, instead learning implicit priors through deep architectures trained on paired data.

Despite their success, these approaches share a fundamental limitation: inaccurate or inadequate priors lead to uncontrollable reconstruction trajectories. When the prior term fails to accurately model the true data distribution, the iterative optimization process may follow erratic or inefficient paths through the solution space. This trajectory instability manifests differently across methods. In PnP approaches, MMSE denoisers often over-smooth images, causing the optimization to take circuitous paths that heavily depend on manual parameter tuning for convergence. Deep unfolding networks exhibit training instability due to their sensitivity to initialization and training data. These trajectory deviations not only slow convergence but also compromise reconstruction quality.

Therefore, we argue for a fundamental paradigm shift: rather than letting priors implicitly determine trajectories, we propose to jointly design both the reconstruction trajectory and the prior network. This dual approach enables explicit control over how the optimization evolves while simultaneously learning more effective priors. Our work is inspired by the trajectory design in generative models such as Flow Matching~\cite{lipman2022flow}, InDi~\cite{delbracio2023inversion} and diffusion bridge ~\cite{chung2023direct}, where they share a similar trajectory spirit: the intermediate states is interpolated by taking a convex combination of initial and last states. By learning to interpolate between these intermediate states, we transform the abrupt measurement-to-image mapping into a smooth, controllable flow that naturally avoids local minima and training instabilities.

Building on these insights, we design a trajectory-controllable unfolding network specifically tailored for hyperspectral reconstruction. Our approach introduces learnable trajectory interpolation mechanisms that enable continuous control over the reconstruction path, transforming the discrete unfolding process into a flexible continuous trajectory. Through our analysis, we discover that skip connections play a crucial role in shaping this trajectory by modulating information flow across iterations. This motivates us to design a frequency-aware skip connection module that adaptively controls the propagation of different frequency components throughout the unfolding process. Furthermore, recognizing the unique characteristics of hyperspectral data, we develop an efficient spatial-spectral transformer architecture that captures long-range dependencies in both spatial and spectral dimensions. 

Our main contributions are summarized as follows:
\begin{itemize}
\item We propose a trajectory-controllable compressed spectral reconstruction algorithm based on back-projection unfolding networks, dubbed FLoUNet. By decomposing the ill-posed inverse problem into a series of progressively easier optimization steps, our framework ensures training stability throughout the reconstruction process and improves reconstruction quality compared to traditional unfolding approaches.

\item We design an efficient spatial-spectral hybrid network architecture with a novel frequency-aware fusion module in the U-Net skip connections. Our analysis reveals that skip connections play a crucial role in trajectory control, and our frequency-aware design enables better information flow across different spectral components.

\item Extensive experiments demonstrate that our method achieves state-of-the-art performance on both simulated and real-world hyperspectral datasets, validating the effectiveness of our trajectory-controllable framework and spatial-spectral network design.
\end{itemize}

\section{Related Work}

\subsection{Hyperspectral Image Reconstruction}

To recover the inverse problem in CASSI, one often use proximal algorithms to solve the optimization problems of form ~\eqref{eq:map_formulation} when the prior functions are non-smooth, by leveraging the proximal operator,
\begin{equation}
\text{prox}_{\lambda h}(\mathbf{v}) = \arg\min_{\mathbf{x}} \left\{\frac{1}{2\lambda}\|\mathbf{x} - \mathbf{v}\|_2^2 + h(\mathbf{x})\right\}
\label{eq:proximal_operator}
\end{equation}
where $h(\mathbf{x}) = -\log p(\mathbf{x})$ and $\lambda$ is a regularization parameter. The PnP concept suggests replacing the proximal operator applying as an off-the-shelf Gaussian denoiser, whereas deep unfolding scheme treats the proximal operator as a trainable neural network. In either scheme, we can formulate the typical iterative update as:
\begin{align}
\mathbf{z}_k &\leftarrow \mathbf{x}_k - \gamma \frac{\partial}{\partial \mathbf{x}_k} \left\| \mathbf{W} \left( \mathbf{y} - \mathbf{H} \mathbf{x}_k \right) \right\|_2^2, \label{eq:weighted_gradient_step} \\
\mathbf{x}_{k+1} &\leftarrow \text{prox}_{\sigma}(\mathbf{z}_k), \label{eq:denoising_step}
\end{align}
where we use a generalized weighted gradient step ~\eqref{eq:weighted_gradient_step} to signify the measurement consistency updating and a network $\text{prox}_{\sigma}$ to denote the proximal operator, $\gamma$ is the step size and $\mathbf{W}$ depends on the noise covariance, if the noise is Gaussian~\cite{chung2022improving}. In PnP scheme, a pretrained MMSE denoiser $\mathbb{E}[\mathbf{x}\mid\mathbf{x_{k}}]$ is used to approximate $\text{prox}_{\sigma}(\mathbf{z}_k)$ whereas a learnable network $\mathcal{D}_{\theta}$ (with parameter ${\theta}$) is used in unfolding framework. \cite{zheng2021deep} employs a spectral denoiser as a PnP prior for CASSI reconstruction, but it suffers from suboptimal reconstruction quality and relies heavily on manual parameter tuning. In comparison, deep unfolding networks enable end-to-end training of both data fidelity and prior components, while maintaining interpretability grounded in optimization algorithms. Recent efforts have focused on enhancing network architectures to better capture the intrinsic structures of hyperspectral data. For instance, \cite{cai2022mask} and \cite{dong2023residual} introduce spectral attention mechanisms to model inter-spectral correlations. In contrast, \cite{cai2022degradation} and \cite{zhang2024dual} leverage spatial attention to extract both local and non-local spatial features of hyperspectral images (HSIs). Furthermore, \cite{li2023pixel} proposes a hybrid spatial-spectral transformer to jointly model spatial and spectral dependencies.

\begin{figure}[t]
    \centering
    \vspace{-2mm}
    \includegraphics[width=0.48\textwidth]{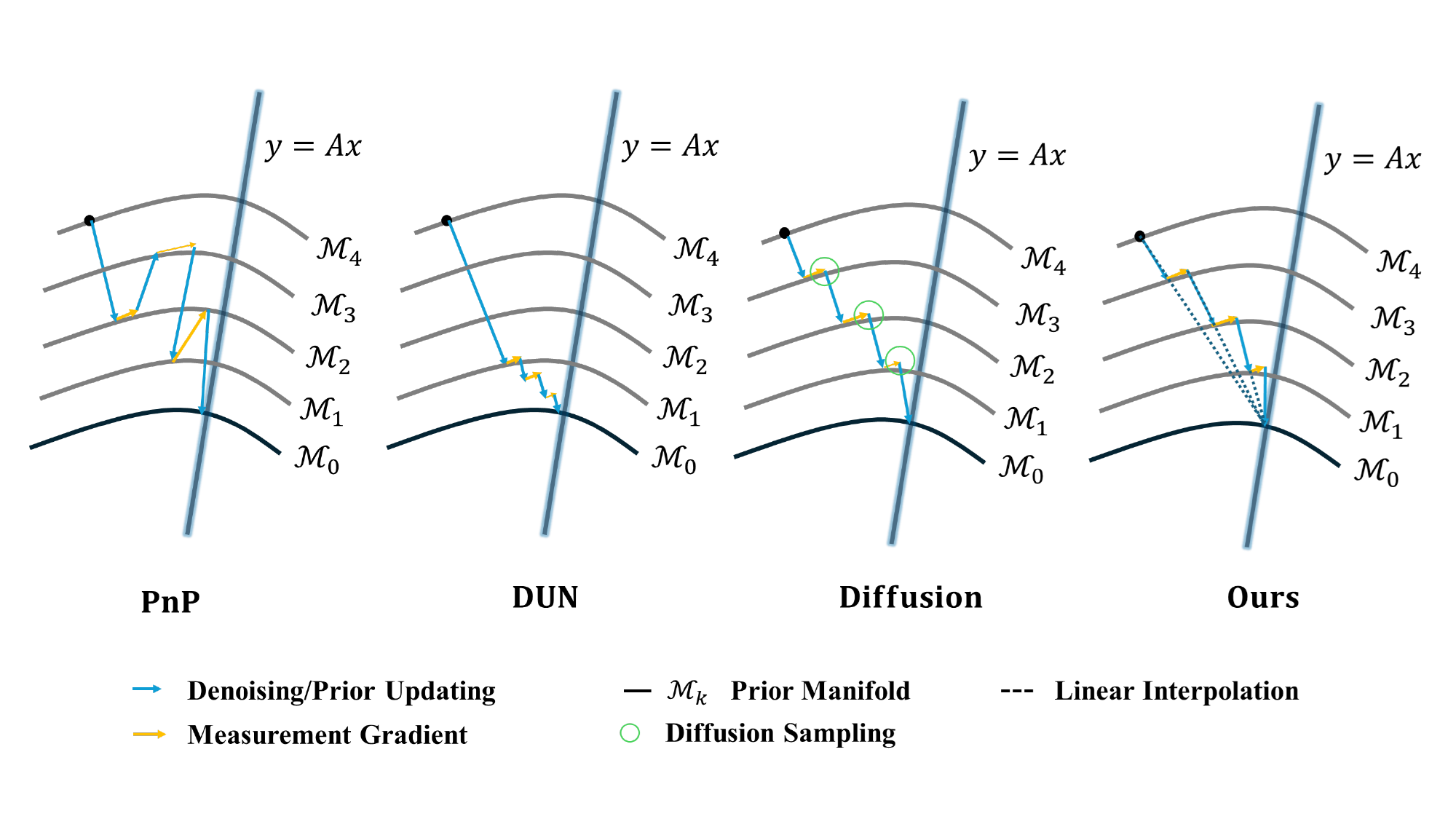}
    \vspace{-5mm}
    \caption{\small  Iterative trajectory of model-based methods for inverse problem. Detail discussion will be in the Appendix. }
    \label{fig:main}
     \vspace{-2mm}
\end{figure}


\subsection{Diffusion Models for Inverse Problems}
Diffusion models (DMs) have emerged as powerful generative priors for inverse problems, using iterative SDE-based sampling guided by learned score functions to gradually refine noise into a clean signal. The posterior score (gradient of the log posterior) can be estimated by combining the prior and likelihood terms; for example, in an unconditional DM one can inject the observation $y$ at inference via Bayes’ rule, $\nabla_x \log p(x|y) = \nabla_x \log p(x) + \nabla_x \log p(y|x)$, whereas conditional DMs are trained to model $p(x|y)$ directly. Recent works explore different paradigms for integrating DMs into inverse problems. Flow Matching ~\cite{lipman2022flow} replaces the stochastic diffusion process with a continuous normalizing flow, learning an ODE that transports a simple distribution (e.g. Gaussian noise) into the data distribution along a fixed path. This simulation-free approach (an ODE-based CNF) subsumes traditional diffusion as a special case and can incorporate inverse-problem constraints by injecting measurement-informed score corrections (analogous to posterior sampling adjustments). In contrast, InDi (Inversion by Direct Iteration) ~\cite{delbracio2023inversion} forgoes the explicit noise-injection paradigm and learns a direct iterative restoration mapping: starting from the degraded input, a neural network applies a sequence of small refinements to recover the clean image. Trained on paired low/high-quality data, InDi avoids the “regression to the mean” effect of one-step predictions and produces more realistic, detailed reconstructions. Meanwhile, Diffusion Bridges ~\cite{chung2023direct} define a diffusion process that directly connects the corrupted (measured) distribution to the clean distribution. Recently, several works have extended DMs to the CASSI hyperspectral reconstruction task.  LADE-DUN ~\cite{wu2024latent} integrates a latent diffusion model into a deep unfolding network, injecting prior spectral knowledge to guide each reconstruction stage. PSR-SCI \cite{Zeng_2025_ICLR} reformulates MSI recovery as subspace unmixing and refines it via diffusion in a low-dimensional space. These approaches demonstrate the synergy between generative priors and physics-constrained reconstruction in spectral imaging.

\section{Methodology}

\subsection{Reinterpreting the CASSI Forward Model}
Traditional CASSI reconstruction methods represent the entire measurement process as a single matrix $\mathbf{A}$, which obscures the underlying physics and system structure. In this section, we decompose the forward model to explicitly represent the spatial modulation and spectral dispersion operations. The CASSI system captures a 3D hyperspectral cube $\mathbf{X} \in \mathbb{R}^{H \times W \times L}$ in a single 2D measurement $\mathbf{Y} \in \mathbb{R}^{H \times (W+L-1)}$, where $H$ and $W$ denote the spatial dimensions and $L$ represents the number of spectral bands. The imaging process consists of three key components: (1) a coded aperture (mask) that spatially modulates the incident light, (2) a dispersive element (prism or grating) that introduces wavelength-dependent lateral shifts, and (3) a detector that integrates the superimposed spectral information.

We reformulate the CASSI forward model by introducing permutation matrices to explicitly model the spectral dispersion. Let $\mathbf{M} \in {0,1}^{H \times W}$ denote the binary coded aperture pattern, $\mathbf{X}_i \in \mathbb{R}^{H \times W}$ represent the $i$-th spectral band of the hyperspectral cube, and $\text{vec}(\cdot)$ dennotes the vectorized form of matrix. The vectrized spectra is expressed as $\mathbf{x}_i=\textbf{vec}( \mathbf{X}_i)$. The forward process can be expressed as:
\begin{equation}
\mathbf{y} = \sum_{i=1}^{L} \mathbf{P}_i \operatorname{diag}(\mathbf{m}) \mathbf{x}_i + \mathbf{n}, 
\label{eq:cassi_permutation}
\end{equation}
where $\operatorname{diag}(\cdot)$ denotes a diagonal matrix formed from a vector, $\mathbf{m} = \operatorname{vec}(\mathbf{M})$ is the vectorized binary coded aperture pattern, $\mathbf{n} \in \mathbb{R}^{HW}$ is the measurement noise, and $\mathbf{P}_i \in \{0,1\}^{HW \times HW}$ is a permutation matrix modeling the wavelength-dependent lateral shift for the $i$-th spectral band. While the (block) composite operator $\mathbf{P}_i \operatorname{diag}(\mathbf{m})$ can serve directly as the forward matrix, its structure poses difficulties for efficient inverse computations (e.g., evaluating $\mathbf{A}\mathbf{A}^\top$). To address this, we reformulate the model using an equivalent yet more computationally friendly form. Firstly we introduce the following theorem.

\vspace{0.5em}
\noindent
\textbf{Theorem 1.} \textit{
Let $\mathbf{v} \in \mathbb{R}^{n}$ be an arbitrary vector and $\mathbf{P} \in \{0,1\}^{n \times n}$ be a permutation matrix. Then, the following identity holds:
}
\begin{equation}
\operatorname{diag}(\mathbf{v}) \cdot \mathbf{P} = \mathbf{P} \cdot \operatorname{diag}(\mathbf{P}^\top \mathbf{v}).
\label{eq:perm_identity}
\end{equation}

\vspace{0.5em}
\noindent
This identity enables us to transform the forward model in Eq.~\eqref{eq:cassi_permutation} into a more physically meaningful form:
\begin{equation}
\mathbf{y} = \sum_{i=1}^{L} \operatorname{diag}(\mathbf{P}_i \mathbf{m}) \cdot \mathbf{P}_i \mathbf{x}_i + \mathbf{n},
\label{eq:cassi_shifted}
\end{equation}
where each spectral band $\mathbf{x}_i$ is first spatially shifted by $\mathbf{P}_i$, then modulated by the shifted mask $\mathbf{P}_i \mathbf{m}$, and finally integrated by the detector. The proof of Theorem 1 is provided in the supplementary material.

Let us denote $\mathbf{x}^{\textbf{shift}}_i = \mathbf{P}_i\mathbf{x}_i$ as the shifted version of the $i$-th spectral band and $\mathbf{M}^{\textbf{shift}}_i = \operatorname{diag}(\mathbf{P}_i\mathbf{m})$ as the corresponding shifted mask pattern. The measurement can then be rewritten as:
\begin{equation}
\mathbf{y} = \sum_{i=1}^{L} \mathbf{M}^{\textbf{shift}}_i \odot \mathbf{x}^{\textbf{shift}}_i + \mathbf{n}.
\label{eq:cassi_shifted}
\end{equation}
The equation ~\eqref{eq:cassi_shifted} is equivalent to a linear forward model that can be expressed as:
\begin{equation}
\mathbf{y} = \mathbf{A}\mathbf{x}^{\textbf{shift}}_i + \mathbf{n}
\label{eq:cassi_compact}
\end{equation}
This formulation reveals that CASSI effectively applies different mask patterns $\mathbf{M}'_i$ to each shifted spectral band $\mathbf{x}^{\textbf{shift}}_i$. The permutation operation $\mathbf{P}_i$ induces a lateral shift of $(i-1)$ pixels for the $i$-th band, creating a dispersed superposition on the detector plane. Crucially, the permutation matrices $\mathbf{P}_i$ are orthogonal, satisfying $\mathbf{P}_i^\top\mathbf{P}_i = \mathbf{P}_i\mathbf{P}_i^\top = \mathbf{I}$. This orthogonality property ensures that the shifting operation is fully reversible. Once we reconstruct the shifted spectral bands $\mathbf{x}^{\textbf{shift}}_i$ from the compressed measurement, we can recover the original aligned hyperspectral cube through the inverse permutation:
\begin{equation}
\mathbf{x}_i = \mathbf{P}_i^\top \mathbf{x}^{\textbf{shift}}_i.
\label{eq:inverse_permutation}
\end{equation}
This vectorized formulation facilitates the integration with optimization algorithms and deep learning frameworks.

\begin{figure}[t]
    \centering
    \vspace{-2mm}
    \includegraphics[width=0.48\textwidth]{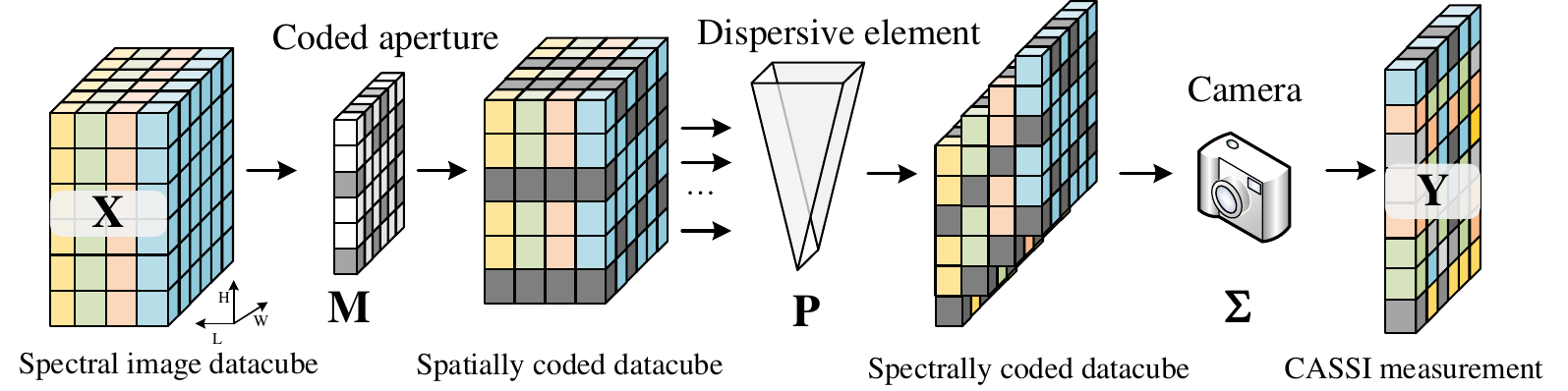}
    \vspace{-5mm}
    \caption{\small  Forward model of CASSI.}
    \label{fig:main}
     \vspace{-2mm}
\end{figure}

\subsection{Back-projection unfolding framework}

Unfolding networks simulate iterative optimization algorithms through a finite number of stages, each consisting of two main components: a data-fidelity update~\eqref{eq:weighted_gradient_step} and a prior-denoising step~\eqref{eq:denoising_step}. In our framework, we adopt a \textit{back-projection (BP)} strategy~\cite{garber2024image} to enforce measurement consistency. Specifically, after each denoising operation, we project the estimate back to the measurement-consistent space by solving the following constrained least-squares problem:
\begin{equation}
\mathbf{x}_{k+1} = \arg\min_{\mathbf{x}} \| \mathbf{x} - \mathbf{x}_{k} \|_2^2 \quad \text{s.t.} \quad \mathbf{A}\mathbf{x} = \mathbf{y},
\end{equation}
with the closed-form solution
\begin{equation}
\mathbf{x}_{k+1} = \mathbf{x}_{k} - \mathbf{A}^{\dagger} (\mathbf{A}\mathbf{x}_{k} - \mathbf{y}),
\label{prac_datafidelity}
\end{equation}
where $\mathbf{x}_{k}$ is the denoised estimate at stage $k$, and $\mathbf{A}^{\dagger}$ is the Moore–Penrose pseudoinverse. To improve stability in underdetermined systems such as CASSI, we use a regularized approximation:
\begin{equation}
\mathbf{A}^{\dagger} \approx \mathbf{A}^\top (\mathbf{A}\mathbf{A}^\top + \eta \mathbf{I})^{-1},
\end{equation}
where $\eta > 0$ is a regularization constant. According to~\cite{cai2022degradation}, the matrix $\mathbf{A}\mathbf{A}^\top$ is diagonal in CASSI, making this projection step computationally efficient and scalable.

To model the image prior, we employ a lightweight spatial-spectral transformer, which will be detailed in the next section. Beyond architectural design, we also draw inspiration from recent advances in generative modeling and trajectory optimization~\cite{chung2023direct, delbracio2023inversion}. These works reveal that intermediate states along the reverse-time denoising path can be interpreted as optimal interpolants between the clean signal and noisy measurements. To leverage this insight, we introduce a theoretical formulation that enables stage-wise supervision over intermediate reconstructions. Assume that the denoising network approximates the MMSE estimator: $\mathbf{x}_{k} \approx \mathbb{E}[\mathbf{x}\mid\mathbf{x}_k]$. Then, the following proposition holds.

\vspace{0.5em}
\noindent
\textbf{Proposition 1.} \textit{
Let $x_s, x_t$ be intermediate stages where $s \geq t$. If the denoiser is an MMSE estimator, then:
}
\begin{equation}
\mathbb{E}[x_s \mid x_k] = \left(1 - \frac{s}{k} \right)\mathbb{E}[x\mid x_k] + \frac{s}{t} x_k.
\label{eq:conditional_expectation}
\end{equation}
According to Proposition~\ref{eq:conditional_expectation}, the next-stage update can be interpreted as a convex combination between the current noisy variable $x_k$ and its denoised estimate $\mathbb{E}[x_s \mid x_k]$. Formally, the unfolding update rule becomes:
\begin{equation}
x_{k+1} = \left(1 - \frac{s}{K} \right) \mathcal{D}_\sigma(x_k) + \frac{s}{K} x_k,
\label{eq:stage_update}
\end{equation}
where $\mathcal{D}_\sigma(x_k)$ denotes the output of the denoising prior at stage $k$, and $K$ is the total number of unfolding stages. In practice the weighting parameter $\frac{s}{K}$ is learned during training. This identity provides a principled way to interpolate denoising trajectories, facilitating smoother signal transitions. To encourage consistency with this trajectory, we introduce a stage-wise supervision loss during training:
\begin{equation}
\mathcal{L}_{\text{traj}} = \sum_{k=1}^{K} \alpha(k, K) \cdot \left\| x_k - \left( \left(1 - \frac{s}{K} \right)x + \frac{s}{K} x_k \right) \right\|_2^2,
\end{equation}
which ensures that intermediate reconstructions are well-aligned with the theoretical reverse-time path. $\alpha(s, t)$ is a weighting parameter to account for the varying importance of intermediate stages, which is defined as:
\begin{equation}
\alpha(s, t) = 1 - \exp\left(-k \cdot \frac{s}{t} \right), \quad \text{where } k > 0.
\end{equation}
An alternating update between ~\eqref{prac_datafidelity} and ~\eqref{eq:stage_update} lead to a gradual optimization toward clean spectral images. Interestingly, we find this linear trajectory shares the same principle in gradient step denoiser~\cite{hurault2021gradient} and proximal denoiser~\cite{wang2025proximal} with a guaranteed fixed-point convergence. The theoretical derivation is provided in the supplementary material.

\begin{figure*}[!htb]
  \centering
  \includegraphics[width=0.8\linewidth]{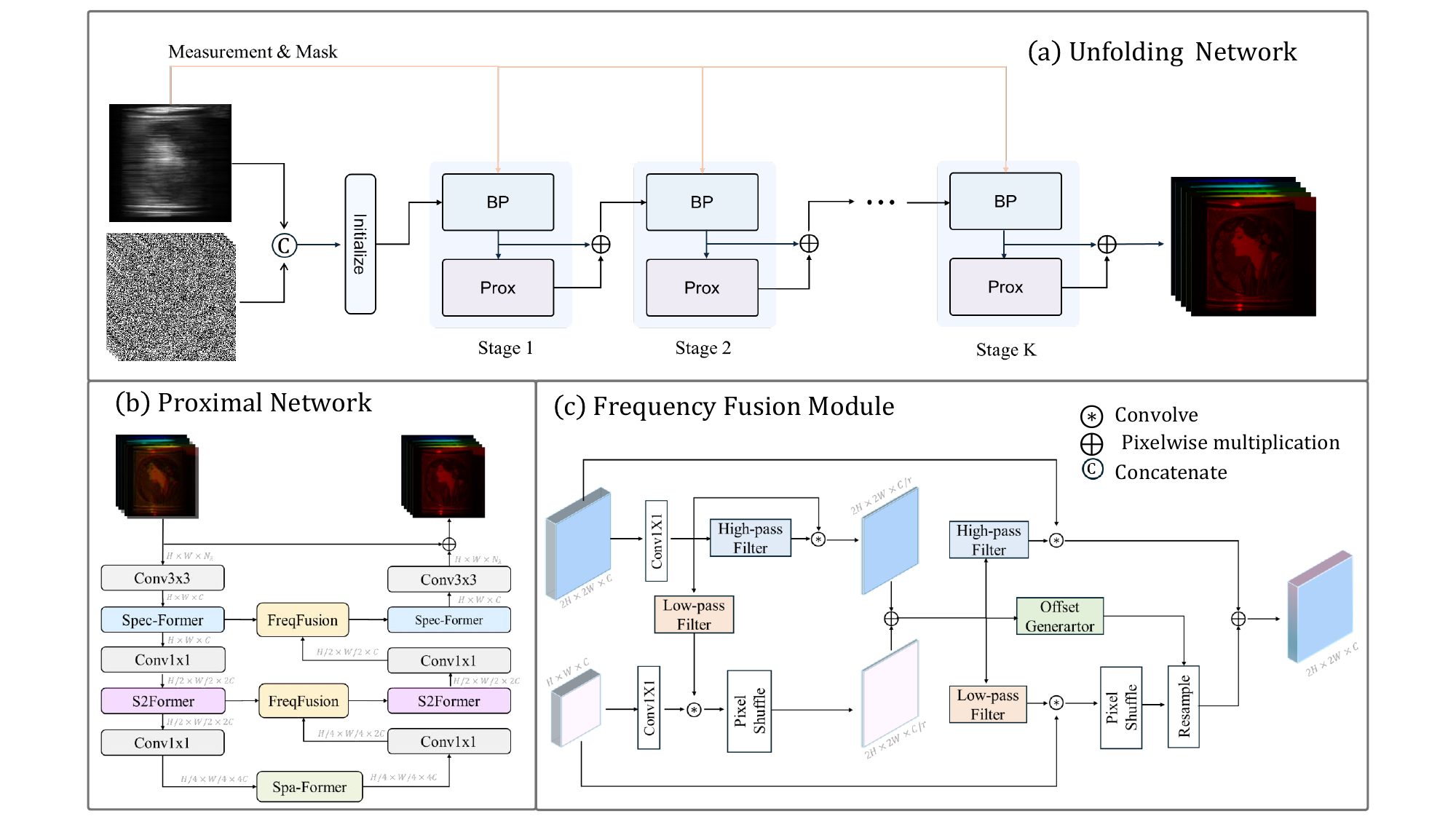}
  \caption{An overview of our proposed unfolding network for HSI reconstruction task, including unfolding pipeline, proximal network and frequency-domain fusion.}
  \label{fig:algo_overview}
\end{figure*}

\subsection{Proximal network}
The proximal network estimates the clean signal prior at each unfolding stage. To ensure efficient hyperspectral reconstruction, we design a hybrid attention strategy that leverages spectral attention in shallow layers for enhanced channel modeling, and spatial attention in deeper layers to capture contextual dependencies. This complements the hierarchical structure of U-Net and balances accuracy with efficiency.

To further refine skip connections, we introduce a frequency-aware fusion module that transforms encoder and decoder features into the frequency domain. By selectively merging high-frequency textures and low-frequency semantics, the module enhances detail and alignment before reprojecting to the spatial domain.

Overall, our proximal network integrates spatial-spectral attention and frequency-domain fusion to provide a compact and effective prior tailored for hyperspectral reconstruction.

\begin{figure*}[!htb]
  \centering
  \includegraphics[width=0.95\linewidth]{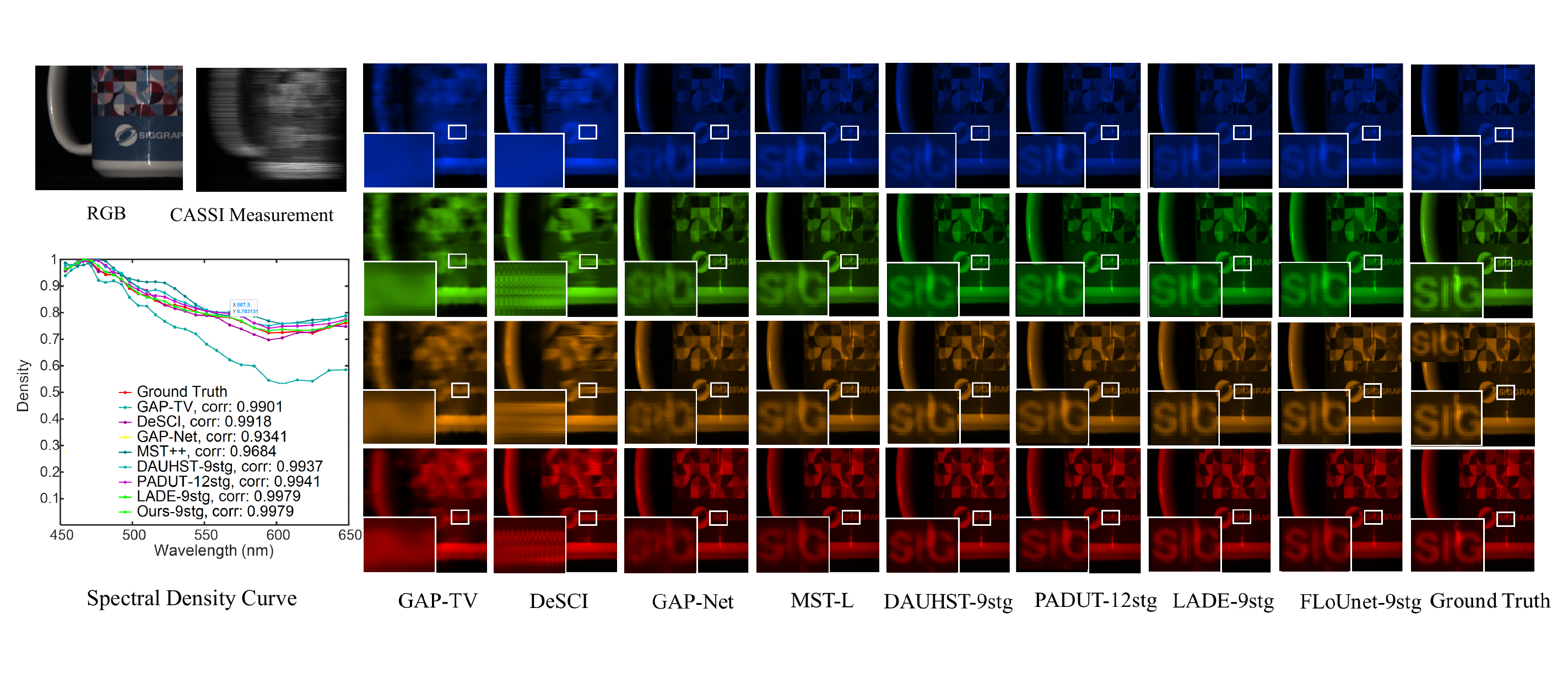}
  \caption{An overview of our proposed unfolding network for HSI reconstruction task, including unfolding pipeline, proximal network and frequency-domain fusion.}
  \label{fig:simu_results}
   \vspace{4mm}
\end{figure*}

\noindent{\bf Efficient Spatial-Spectral Transformer}

\noindent
Let the input feature in each local window be denoted as $X \in \mathbb{R}^{L^2 \times C}$, where $C$ is the number of spectral channels and $L^2$ is the number of spatial tokens in the window. We first apply layer normalization and compute the query, key, and value:
\begin{equation}
Q\!=\!\text{Conv}_{1\!\times\!1}\!(X),\, K\!=\!\text{Conv}_{1\!\times\!1}\!(X),\, V\!=\!\text{Conv}_{1\!\times\!1}\!(X)
\end{equation}
where $Q, K, V \in \mathbb{R}^{L^2 \times C}$. For spectral attention, we treat each channel as a token and perform attention across channels. This is realized by transposing $Q$ and $K$ and apply matrix multiplication:
\begin{align}
\text{Attn}_{\text{spec}} &= \text{Softmax}\left(\frac{Q^\top K}{\sqrt{L^2}}\right) \in \mathbb{R}^{C \times C}, \\
Y &= \text{Attn}_{\text{spec}} \cdot V^\top \in \mathbb{R}^{C \times L^2}
\end{align}
We then apply a projection to the original shape by $X' = \text{Conv}_{1 \times 1}(Y^\top) \in \mathbb{R}^{L^2 \times C}$.

Similarly, for spatial attention, we use low-rank projections to reduce the attention cost. As the feature channel increases, we project the query and key to a low-rank space $C'$:

\begin{equation}
\small
Q = \text{Conv}_{1 \times 1}(X), 
K = \text{Conv}_{1 \times 1}(X), 
V = \text{Conv}_{1 \times 1}(X)
\end{equation}
where $Q, K \in \mathbb{R}^{L^2 \times C'}$, $V \in \mathbb{R}^{L^2 \times C}$, and $C' < C$ is the reduced dimension for low-rank projection.
Spatial attention is then computed across spatial positions:
\begin{align}
\text{Attention}_{\text{spa}} &= \text{Softmax}\left(\frac{Q K^\top}{\sqrt{C'}}\right) \in \mathbb{R}^{L^2 \times L^2}, \\
Y &= \text{Attention}_{\text{spa}} \cdot V \in \mathbb{R}^{L^2 \times C}, \nonumber \\
X' &= \text{Conv}_{1 \times 1}(Y).
\end{align}

\vspace{0.5em}
\noindent
Overall, the use of hybrid spectral and spatial attention allows our transformer to model local high-frequency structures and long-range dependencies in a computationally tractable way. 



\noindent{\bf Frequency-Aware Feature Fusion}

\noindent To enhance the quality of feature integration in our network, we incorporate a frequency-aware feature fusion mechanism. This module aims to balance semantic and structural information by performing both initial and final fusion operations. It consists of three key components: the Low-Pass Filter (LPF), the High-Pass Filter (HPF), and the Offset Generator, as illustrated in Appendix. The proposed
FreqFusion can be formally written as:
\begin{align}
\hat{\mathbf{F}}_{dec}^{t+1} &= \hat{\mathbf{F}}_{enc}^{t} + \hat{\mathbf{F}}_{dec}^{t}, \label{eq:initial_fusion_1} \\
\hat{\mathbf{F}}_{dec}^{t} &= \mathcal{F}^{\text{UP}}\left(\mathcal{F}^{\text{LP}}(\mathbf{F}_{dec}^{t})\right), \label{eq:initial_fusion_2} \\
\hat{\mathbf{F}}_{enc}^{t} &= \mathcal{F}^{\text{HP}}(\mathbf{F}_{enc}^{t}) + \mathbf{F}_{enc}^{t}. \label{eq:initial_fusion_3}
\end{align}
Before performing frequency-aware fusion, we first align low-resolution features $\mathbf{F}_{dec}^{k}$ from Decoder to the same channel size with high-resolution features $\mathbf{F}_{enc}^{k}$ from Encoder by $\mathbf{F}_{dec}^{k} = \text{Conv}_{1\times1}(\mathbf{F}_{dec}^{k})$.

\paragraph{Low-Pass Filter.}
The LPF generator predicts dynamic, spatially-varying low-pass filters:
\begin{equation}
\mathbf{Y}^{\text{LP}} = \mathcal{F}^{\text{LP}}(\mathbf{F}_{dec}^{t})
\end{equation}
These filters smooth intra-object variations, enhancing feature consistency while preserving coarse spatial structures.

\paragraph{High-Pass Filter.}
To retain high-frequency details such as edges and textures, the HPF generator performs:
\begin{equation}
\mathbf{Y}^{\text{HP}} = \mathcal{F}^{\text{HP}}(\mathbf{F}_{enc}^{t}) + \mathbf{F}_{enc}^{t}
\end{equation}
where $\mathcal{F}^{\text{HP}}$ extracts high-frequency signals by subtracting blurred outputs (average-pooled results) from learned filters. This strengthens boundary fidelity in the fused features.

\paragraph{Offset Generator.}
The offset generator adaptively predicts spatial offsets based on local cosine similarity, enabling resampling toward high intra-category similarity regions to correct inconsistent features and refine object boundaries. This is to be discussed in appendix.

\section{Experiments}

\begin{table*}[ht!]
\begin{center}
  \resizebox{0.98\textwidth}{!}
   {
  \begin{tabular}{c|cc|cccccccccc|c}
    \toprule
Algorithms&Params(M) &FLOPs(G) & Scene1 & Scene2 & Scene3 & Scene4 & Scene5 & Scene6 & Scene7 & Scene8 & Scene9 & Scene10 & Avg \\ 
\midrule

 && & 26.82&22.89&26.31&30.65&23.64&21.85&23.76&21.98&22.63&23.1&24.36\\
 \multirow{-2}{*}{GAP-TV~\cite{yuan2016generalized}}   &  \multirow{-2}*{---} & \multirow{-2}*{---}   & 0.754&0.610&0.802&0.852&0.703&0.663&0.688&0.655&0.682&0.584&0.669\\ 
\midrule

  && & 27.13 & 23.04  & 26.62& 34.96& 23.94& 22.38& 24.45 & 22.03& 24.56& 23.59& 25.27 \\
 \multirow{-2}{*}{DeSCI~\cite{liu2018rank}}   &  \multirow{-2}*{---} & \multirow{-2}*{---}   & 0.748& 0.620 & 0.818& 0.897& 0.706& 0.683& 0.743  & 0.673& 0.732 & 0.587 &0.721  \\ 
\midrule

 && &  33.63 & 33.19& 33.96& 39.14& 31.44& 32.29& 31.79& 30.25& 33.06& 30.14& 32.89\\
 \multirow{-2}{*}{GAP-Net~\cite{meng2023deep}}& \multirow{-2}*{4.27} &  \multirow{-2}*{78.58} &  0.913 & 0.902& 0.931& 0.971& 0.921& 0.927& 0.903& 0.907& 0.916& 0.898& 0.919 \\ 
\midrule 

 &  &   & 34.96 &35.64&35.55&41.64&32.56&34.33&33.27&32.26&34.17&32.22&34.66 \\
 \multirow{-2}{*}{HDNet~\cite{hu2022hdnet}}&\multirow{-2}*{2.37}  &  \multirow{-2}*{154.76}        & 0.937&0.943&0.940&0.976&0.948&0.950&0.920&0.945&0.944&0.940&0.946\\ 
\midrule

  &  &  &36.78&37.89&40.61&46.93&35.42&35.30&36.58&33.95&39.46&32.80&37.57\\
 \multirow{-2}{*}{BIRNAT~\cite{cheng2022recurrent}}     & \multirow{-2}*{4.40} &  \multirow{-2}*{212.55}  &0.951&0.957&0.971&0.985&0.963&0.959&0.954&0.955&0.969&0.937&0.960\\ 
\midrule

&  &  & 35.96& 36.84& 38.16& 42.44& 33.25& 35.72& 34.86& 34.34 & 36.51 & 33.09& 36.12\\
 \multirow{-2}{*}{CST-L+~\cite{cai2022coarse}}&\multirow{-2}*{3.00}  & \multirow{-2}*{40.10}           & 0.949& 0.955& 0.962& 0.975& 0.955& 0.963& 0.944& 0.961& 0.957& 0.945& 0.957 \\ 
\midrule 

 & & & 35.57&36.22&37.00&42.86&33.27&35.27&34.05&33.50&36.17&33.26&35.72\\
 \multirow{-2}{*}{MST++~\cite{cai2022mask}}  &  \multirow{-2}*{1.33} & \multirow{-2}*{19.42}& 0.945 &0.949&0.959&0.980&0.954&0.960&0.936&0.956&0.956&0.949&0.955 \\ 
\midrule

 & && 37.25& 39.02& 41.05& 46.15& 35.80& 37.08& 37.57 & 35.10& 40.02& 34.59& 38.36 \\
 \multirow{-2}{*}{DAUHST-9stg~\cite{cai2022degradation}}   & \multirow{-2}*{6.15} &   \multirow{-2}*{79.50} & 0.958& 0.967& 0.971& 0.983& 0.969& 0.970& 0.963 & 0.966  & 0.970  & 0.956& 0.967  \\ 
\midrule

 &  & &37.36 &40.43 &42.38 &46.62 &36.26 &37.27 &37.83 &35.33 &40.86 &34.55 &38.89 \\
 \multirow{-2}{*}{PADUT-12stg~\cite{li2023pixel}}  &  \multirow{-2}*{5.38} &  \multirow{-2}*{90.46} &  0.962 &0.978 &0.979 &0.990 &0.974 &0.974 &0.966 &0.974 &0.978 &0.963 &0.974 \\ 
\midrule

 &  & & 37.94& 40.95& 43.25 & 47.83 & 37.11 & 37.47& 38.58 & 35.50  & 41.83 & 35.23 & 39.57 \\
 \multirow{-2}{*}{RDLUF-MixS$^2$-9stg~\cite{dong2023residual}} & \multirow{-2}*{1.89} &  \multirow{-2}*{115.34} & 0.966 & 0.977 & 0.979 & 0.990 & 0.976 & 0.975 & 0.969 & 0.970 & 0.978 & 0.962 & 0.974 \\

\midrule 
 &  & & 37.14 & 39.60 & 41.78 & 46.57 & 35.57 & 37.02 & 36.80 & 35.22 & 40.15 & 34.17 & 38.40 \\
 \multirow{-2}{*}{LADE-3stg~\cite{wu2024latent}} & \multirow{-2}*{1.08} & \multirow{-2}*{36.93} 
 & 0.963 & 0.975 & 0.978 & 0.990 & 0.971 & 0.975 & 0.960 & 0.973 & 0.976 & 0.962 & 0.972 \\

\midrule 
 &  & & 38.08 & 41.84 & 43.77 & 47.99 & 37.97 & 38.30 & 38.82 & 36.15 & 42.53 & 35.48 & 40.09 \\
 \multirow{-2}{*}{LADE-9stg~\cite{wu2024latent}} & \multirow{-2}*{2.78} & \multirow{-2}*{88.68} 
 & 0.969 & 0.982 & 0.983 & 0.993 & 0.980 & 0.980 & 0.973 & 0.979 & 0.984 & 0.970 & 0.979 \\


\midrule
 &  &  & 38.13 & 39.89 & 42.56 & 7.42 & 36.96 & 37.26 & 37.76 & 35.38 & 41.42 & 34.92 & 39.17 \\
\multirow{-2}{*}{FLoUNet-3stg} & \multirow{-2}*{1.35} & \multirow{-2}*{26.18} 
& 0.962 & 0.971 & 0.976 & 0.987 & 0.973 & 0.972 & 0.963 & 0.965 & 0.974 & 0.958 & 0.970 \\

\midrule
 &  &  & 38.76 & 41.18 & 43.87 & 49.18 & 38.27 & 37.77 & 38.78 & 36.39 & 42.82 & 35.71 & 40.27 \\
\multirow{-2}{*}{FLoUNet-9stg} & \multirow{-2}*{4.04} & \multirow{-2}*{78.37} 
& 0.967 & 0.978 & 0.979 & 0.989 & 0.979 & 0.973 & 0.969 & 0.972 & 0.980 & 0.963 & 0.975 \\

 
\bottomrule
\end{tabular}}
\end{center}
\caption{The simulated HSI reconstruction results for Scene 5 with 1 out of 28 spectral channels, including seven state-of-the-art algorithms and our proposed FLoUnet-9stg. The left displays the RGB image and measurement. The bottom-left shows the spectral density curves corresponding to the selected yellow box in the RGB image. }
\label{Table:reordered}
\end{table*}

\subsubsection{Experimental settings}
In the simulation experiments, we use two datasets, CAVE and KAIST. The CAVE dataset comprises 32 HSI images with spatial dimensions of $512 \times 512$. The KAIST dataset contains 30 HSI images, each with spatial dimensions of $2704 \times 3376$. Same as previous researches, we utilize the CAVE dataset as our training set and selected 10 scenes from the KAIST dataset for testing.

\noindent{\bf Implementation Details.}
The proposed model is implemented by Pytorch.
During the training process, we utilize the Adam optimizer ($\beta_1=0.9$ and $\beta_2=0.999$) and a cosine annealing scheduler, running for 300 epochs on a single RTX 4090 GPU.
To evaluate the performance, we use the peak signal-to-noise ratio (PSNR) and structure similarity (SSIM) to assess the HSI reconstruction capabilities.

\subsection{Compare with State-of-the-art}
We compare our proposed FloUnet model with several SOTA CASSI algorithms and the results are analyzed as follows. 
\noindent{\bf Synthetic data.}
To comprehensively evaluate the quantitative results of all competing methods, we test on ten simulated datasets and presented the corresponding numerical results in Tab. \ref{Table:reordered}. Different colors are used to distinguish the types of algorithms: gray for model-based methods, orange for end-to-end networks, and green for deep unfolding methods.
It is evident that our FloUnet model achieved the best numerical results in all cases.
Fig. \ref{fig:simu_results} shows the visual reconstruction results.

\begin{figure}[ht]
  \begin{center}
   \includegraphics[width=0.465\textwidth]{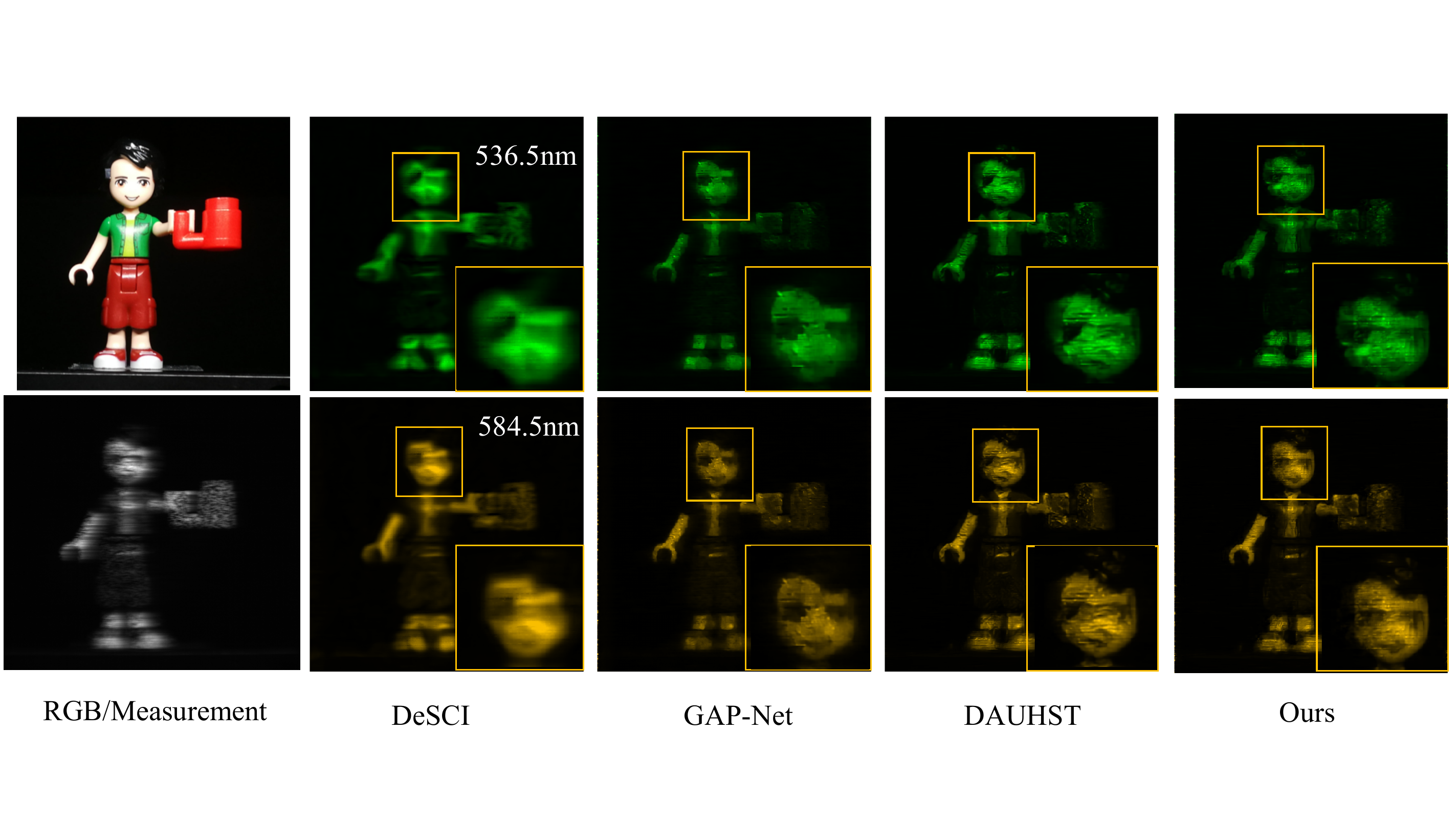}
  \end{center}
\caption{ The real data comparisons.  2 out of 28 wavelengths are plotted for visual comparison. }\label{fig:real}
\end{figure}

\noindent{\bf Real data.}
To further investigate the superiority of this model, we also conduct experiments on real HSI reconstruction tasks. Since the ground truth of real-world scenarios is unattainable, we can only compare qualitative results.
Following the experimental setting of~\cite{cai2022mask}, we apply FLoUnet-9stg to training in the simulated dataset.
Fig. \ref{fig:real}  presents the visual results of our model compared to other algorithms in Scene 4 (2 out of the 28 spectral channels). In comparison, our model can reconstruct more textures and details, but it still exhibits some blurriness and artifacts. These challenges highlight the difficulties the model faces in handling real-world hyperspectral reconstruction tasks.

\begin{table}[t]
\centering
\caption{Ablation study on key components of the proposed method.}
\vspace{-3mm}
\label{tab:ablation_simple}
\scalebox{1}{
\begin{tabular}{lcc}
\toprule
Model Variant & PSNR & SSIM \\
\midrule
Base-1 & 36.77 & 0.949 \\
+ HS2Former & 38.76 & 0.972 \\
+ FreqFusion & 39.01 & 0.970 \\
+ Trajectory Loss & \textbf{39.17} & \textbf{0.971} \\
\bottomrule
\end{tabular}
}
\end{table}

\begin{figure}[t]
    \centering
    \vspace{-2mm}
    \includegraphics[width=0.48\textwidth]{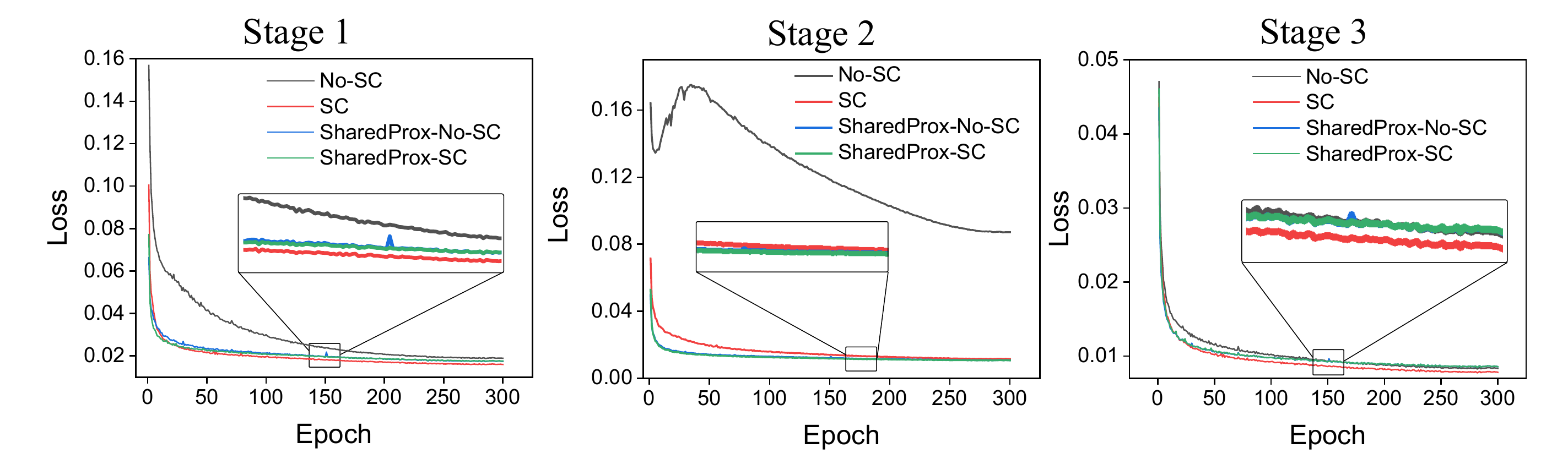}
    \vspace{-5mm}
    \caption{\small  Skip-connection and weight sharing affects the trajectory and convergence in 3 stage unfolding experiment.}
    \label{ Skip-connection}
     \vspace{0mm}
\end{figure}

\subsection{Ablation study}
Our ablation analysis focuses on three main components of FloUnet, hybrid spatial-spectral transformer (we use HS2Former to denote) and Frequency Fusion module and skip-connection. We conduct ablation experiments on public simulated HSI datasets to investigate the effectiveness of each module. 
Tab.~\ref{tab:ablation_simple} summarizes the performance of different cases compared to our model. We choose baseline-1 as the one without spatial-spectral attention with 3 stages. By gradually adding our proposed H2Former, Frequency fusion Module and Trajectory loss, we see the metric PSNR increase around 3dB, which demonstrate the effectiveness of our method. Furthermore, we observed that the skip-connection and weight sharing matters in regularized trajectory optimization of deep unfolding algorithm. It is shown in Figure 6 that by using our hybrid transformer as our backbone (3 stgaes). And we observe the gradual loss decresing on each stages. We observed that without Skip Connection (noted as SC, in between the input and output of Unet structure), the training process become targetless and trajectary is lost (see the 2nd stage). With Skip Connextion, it naturally converge and regularize the stage interaction. The same with weight sharing (noted as SharedProx) in unfolding network, which also guide the unfolding trajectory toward convergence. More ablation study can be found in Appendix.

\section{Conclusion}
In this work, we presented a trajectory-controllable unfolding network tailored for hyperspectral image reconstruction, dubbed FLoUNet. By introducing a learnable trajectory interpolation mechanism, our approach transforms traditional discrete-stage unfolding into a flexible and continuous optimization process, enhancing reconstruction quality and training stability. We further proposed a frequency-aware skip connection module to dynamically regulate the propagation of spatial and spectral information, effectively shaping the optimization trajectory. Additionally, we designed a lightweight spatial-spectral transformer to capture long-range dependencies across both spatial and spectral domains. Extensive experiments on both synthetic and real-world CASSI datasets demonstrate the superiority of our method over existing approaches, highlighting its effectiveness in tackling the challenges of compressive spectral reconstruction through principled trajectory control and efficient hybrid network design.


\bibliography{aaai25}

\section*{Reproducibility Checklist}

\hrule
\vspace{1em}

\checksubsection{General Paper Structure}
\begin{itemize}

\question{Includes a conceptual outline and/or pseudocode description of AI methods introduced}{(yes/partial/no/NA)}
Yes.

\question{Clearly delineates statements that are opinions, hypothesis, and speculation from objective facts and results}{(yes/no)}
Yes.

\question{Provides well-marked pedagogical references for less-familiar readers to gain background necessary to replicate the paper}{(yes/no)}
Yes.

\end{itemize}
\checksubsection{Theoretical Contributions}
\begin{itemize}

\question{Does this paper make theoretical contributions?}{(yes/no)}
Yes.

	\ifyespoints{\vspace{1.2em}If yes, please address the following points:}
        \begin{itemize}
	
	\question{All assumptions and restrictions are stated clearly and formally}{(yes/partial/no)}
	Yes.

	\question{All novel claims are stated formally (e.g., in theorem statements)}{(yes/partial/no)}
	Yes.

	\question{Proofs of all novel claims are included}{(yes/partial/no)}
	Yes.

	\question{Proof sketches or intuitions are given for complex and/or novel results}{(yes/partial/no)}
	Yes.

	\question{Appropriate citations to theoretical tools used are given}{(yes/partial/no)}
	Yes.

	\question{All theoretical claims are demonstrated empirically to hold}{(yes/partial/no/NA)}
	Yes.

	\question{All experimental code used to eliminate or disprove claims is included}{(yes/no/NA)}
	Yes.
	
	\end{itemize}
\end{itemize}

\checksubsection{Dataset Usage}
\begin{itemize}

\question{Does this paper rely on one or more datasets?}{(yes/no)}
Yes.

\ifyespoints{If yes, please address the following points:}
\begin{itemize}

	\question{A motivation is given for why the experiments are conducted on the selected datasets}{(yes/partial/no/NA)}
	Yes.

	\question{All novel datasets introduced in this paper are included in a data appendix}{(yes/partial/no/NA)}
	No.

	\question{All novel datasets introduced in this paper will be made publicly available upon publication of the paper with a license that allows free usage for research purposes}{(yes/partial/no/NA)}
	NA.

	\question{All datasets drawn from the existing literature (potentially including authors' own previously published work) are accompanied by appropriate citations}{(yes/no/NA)}
	Yes.

	\question{All datasets drawn from the existing literature (potentially including authors' own previously published work) are publicly available}{(yes/partial/no/NA)}
	Yes.

	\question{All datasets that are not publicly available are described in detail, with explanation why publicly available alternatives are not scientifically satisficing}{(yes/partial/no/NA)}
	NA

\end{itemize}
\end{itemize}

\checksubsection{Computational Experiments}
\begin{itemize}

\question{Does this paper include computational experiments?}{(yes/no)}
Yes.

\ifyespoints{If yes, please address the following points:}
\begin{itemize}

	\question{This paper states the number and range of values tried per (hyper-) parameter during development of the paper, along with the criterion used for selecting the final parameter setting}{(yes/partial/no/NA)}
	Yes.

	\question{Any code required for pre-processing data is included in the appendix}{(yes/partial/no)}
	No.

	\question{All source code required for conducting and analyzing the experiments is included in a code appendix}{(yes/partial/no)}
	No.

	\question{All source code required for conducting and analyzing the experiments will be made publicly available upon publication of the paper with a license that allows free usage for research purposes}{(yes/partial/no)}
	Yes.
        
	\question{All source code implementing new methods have comments detailing the implementation, with references to the paper where each step comes from}{(yes/partial/no)}
	Yes.

	\question{If an algorithm depends on randomness, then the method used for setting seeds is described in a way sufficient to allow replication of results}{(yes/partial/no/NA)}
	Yes.

	\question{This paper specifies the computing infrastructure used for running experiments (hardware and software), including GPU/CPU models; amount of memory; operating system; names and versions of relevant software libraries and frameworks}{(yes/partial/no)}
	Partial.

	\question{This paper formally describes evaluation metrics used and explains the motivation for choosing these metrics}{(yes/partial/no)}
	Yes.

	\question{This paper states the number of algorithm runs used to compute each reported result}{(yes/no)}
	Yes.

	\question{Analysis of experiments goes beyond single-dimensional summaries of performance (e.g., average; median) to include measures of variation, confidence, or other distributional information}{(yes/no)}
	Yes.

	\question{The significance of any improvement or decrease in performance is judged using appropriate statistical tests (e.g., Wilcoxon signed-rank)}{(yes/partial/no)}
	No.

	\question{This paper lists all final (hyper-)parameters used for each model/algorithm in the paper’s experiments}{(yes/partial/no/NA)}
	Partial.

\end{itemize}
\end{itemize}

\end{document}